\pdfoutput=1

\documentclass[11pt]{article}

\usepackage{acl}

\usepackage{times}
\usepackage{latexsym}

\usepackage[T1]{fontenc}

\usepackage[utf8]{inputenc}
\usepackage{CJK}
\usepackage{enumitem}
\usepackage{microtype}
\usepackage{graphicx}
\usepackage{amsmath}
\usepackage{amsfonts,amssymb}
\usepackage{multirow}
\usepackage{booktabs}
\usepackage{hyperref}
\usepackage{float}

\title{ConFiguRe: Exploring Discourse-level Chinese Figures of Speech}

\author{ \
Dawei Zhu
\quad Qiusi Zhan
\quad Zhejian Zhou
\quad Yifan Song
\\{\bf
Jiebin Zhang
\quad Sujian Li\thanks{\ \ Corresponding author}}
\\
\quad School of Computer Science, Peking University
\\
\quad Key Laboratory of Computational Linguistics, MOE, Peking University\thanks{\ \ Author emails:\{dwzhu, zhanqiusi, zhouzhejian, yfsong, zhangjiebin, lisujian\}@pku.edu.cn }
}

\begin{document}\maketitle

\begin{abstract}
Figures of speech, such as metaphor and irony, are ubiquitous in literature works and colloquial conversations. 
This poses great challenge for natural language understanding since figures of speech usually deviate from their ostensible meanings to express deeper semantic implications.
Previous research lays emphasis on the literary aspect of figures and seldom provide a comprehensive exploration from a view of computational linguistics.
In this paper, we first propose the concept of figurative unit, which is the carrier of a figure.
Then we select 12 types of figures commonly used in Chinese,
and build a Chinese corpus for \underline{Con}textualized \underline{Figu}re \underline{Re}cognition~(ConFiguRe).
Different from previous token-level or sentence-level counterparts, ConFiguRe aims at extracting a figurative unit from discourse-level context, and classifying the figurative unit into the right figure type.
On ConFiguRe, three tasks, i.e.,  figure extraction, figure type classification and figure recognition,  are designed and the state-of-the-art techniques are utilized to implement the benchmarks. We conduct thorough experiments and show that all three tasks  are challenging for existing models, thus requiring further research. Our dataset and code are publicly available at \url{https://github.com/pku-tangent/ConFiguRe}. 

\end{abstract}
\section{Introduction}

Figures of speech, also known as rhetoric figures or figurative languages, are a ubiquitous part of spoken and written discourse. These rhetorical techniques, such as metaphor, irony and parallelism, greatly enrich the expression of human languages~\cite{Roberts:94}.
They intentionally deviate from the literal meaning of language to provide deeper semantic expressiveness, therefore posing a big challenge to natural language understanding.

Linguists have a long history of studying rhetoric figures, extensively analyzing their use in literature, culture and psychology \cite{Zhang63,WILKS197553,paul1998,mip20007,shapin2012ivory}. These works mainly focus on collecting qualitative evidence. Figures of speech have also drawn attention from the NLP community. Many downstream applications could be improved, should figures be precisely identified and carefully dealt with. For example, a faithful translation should adapt the metaphors used in the source language to an authentic expression in the target language, and sentiment analysis should benefit from the correct identification of irony and sarcasm.
\begin{figure}
    \centering
    \includegraphics[scale=0.52]{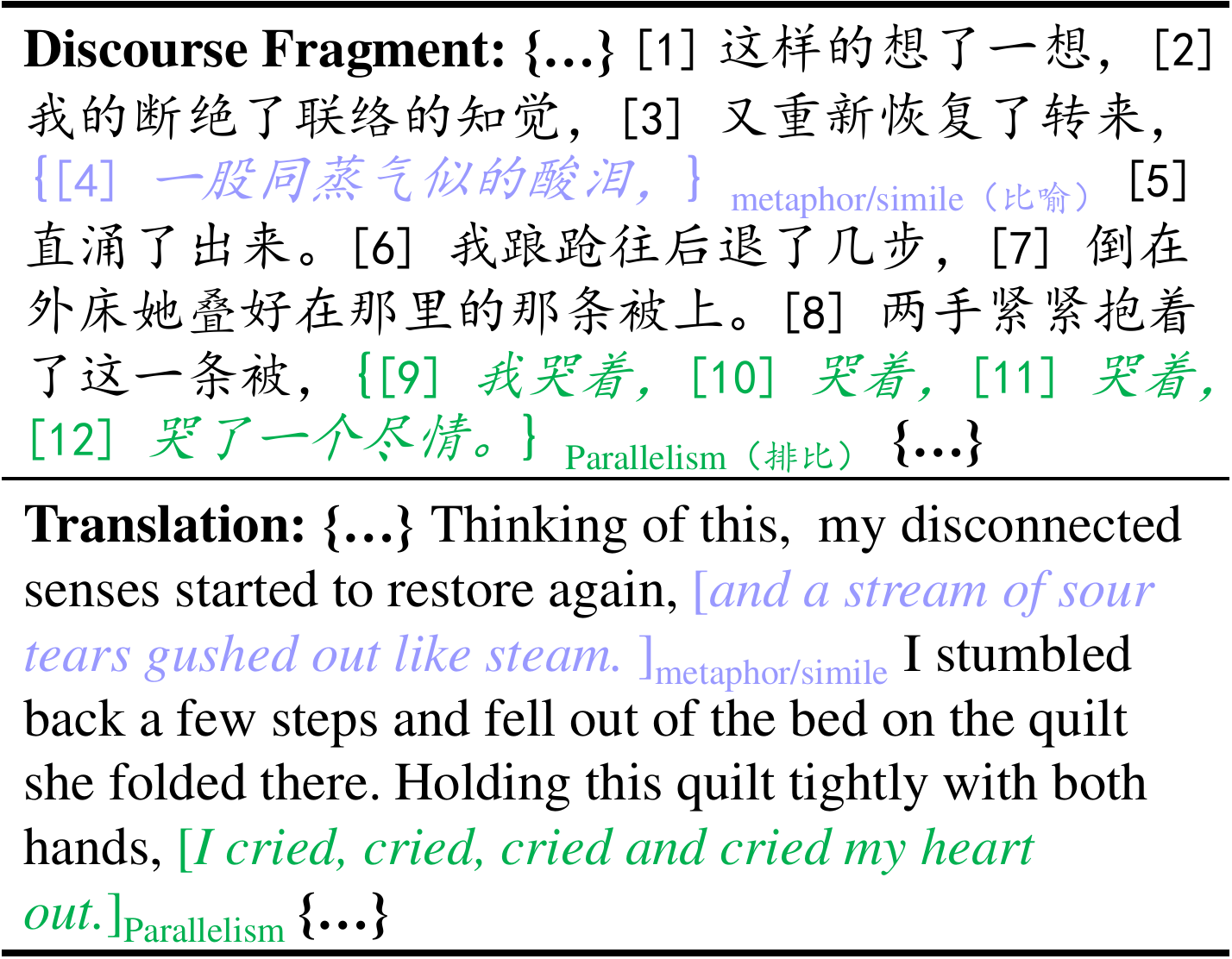}
    \caption{An annotated case of ConFiguRe. We present English translation in the second block for reading convenience. Clause groups in \textit{italic} are figurative units, with the specific figure type denoted in subscription.
    }
    \label{fig:task_example}
\end{figure}

Despite its significance, a comprehensive study of identifying rhetoric figures from discourse remains under-explored.
Previous works mainly emphasize specific figures, such as metaphor~\cite{steen2010a,Fass:91,dodge2015metanet,su-etal-2020-deepmet} and sarcasm \cite{reddit,davidov-etal-2010-semi,wallace-etal-2014-humans,lee-etal-2020-augmenting}, or identify rhetoric figures at a token or sentence level~\cite{rheq_dataset,chen_jointly_2021}.
However, in real-world settings, figurative languages are hidden in a long context and the potential figure type is usually unknown.
Motivated by this, we construct a comprehensive dataset of 12 commonly used figures in Chinese, and include discourse fragment as context for each instance. 

First we describe our guideline towards constructing a rhetoric corpus, which is devised upon linguistic theory and existing reading practice.
The design of the guideline is oriented with two key questions: (1) What is the language carrier of a specific figure? (2) Which figures should be included in our corpus? For the first question, we  propose the concept of \textit{figurative unit}——the smallest continuous
clause sequence containing a complete expression of a specific figure. 
For the second question, we approach it from a linguistic view. Linguistically, figures can be divided into two main groups: \textit{schemes} and  \textit{tropes}. \textit{Schemes} reflect a deviation from the ordinary pattern or arrangement of words, while \textit{tropes} involve deviation from the ordinary and principal signification of words~\cite{corbett1999classical}.
Following previous work of Chinese linguists \cite{Zhang63} and existing reading-comprehension practice, we select 7 tropes and 5 schemes  commonly used in Chinese in our work.

Following the aforementioned guideline, we leverage human annotation to build ConFiguRe, a Chinese corpus for Contextualized Figure Recognition. Each instance in ConFiguRe includes a discourse  fragment with several annotated figurative units attached to it. An annotated instance of ConFiguRe is illustrated in  Figure~\ref{fig:task_example}.  In this instance, the fragment includes 12 Chinese clauses and 2 figurative units. Clause 4  alone is a figurative unit labeled with the \textit{metaphor} type, while the latter 4 clauses constitute another figurative unit labeled \textit{parallelism}. ConFiguRe is, to the best of our knowledge, the first rhetoric dataset that involves both extracting a figurative unit from the discourse-level context and classifying this unit into the right figure type. In comparison, previous datasets mainly focus on detecting a specific figure from a given sentence~\cite{joshi_harnessing_2016,reddit}.

For benchmark settings, we design three tasks based on ConFiguRe, namely \textit{figure extraction}, \textit{figure type classification}, and \textit{figure recognition}. We deploy state-of-the-art models as baselines, and reveal that all three tasks remain challenging through thorough experiment. We also conduct subsidiary experiments to explore future directions for these tasks, which will contribute to the research of this area.

To sum up, our main contribution is threefold:
\begin{itemize}[leftmargin=*]
    \item We design the guideline towards constructing a discourse-level rhetoric corpus based on linguistic theory, and propose the concept of \textit{figurative unit} as the  basic element for analysis.
    \item We construct ConFiguRe, a human-annotated Chinese corpus for contextualized figure recognition, which includes 12 most frequently used figure types. Upon this, we design three tasks as benchmarks: \textit{figure extraction}, \textit{figure type classification}, and \textit{figure recognition}.
    \item We implement models based on recent state-of-the-art techniques as baselines, and conduct thorough experiments and analysis. We find that all three tasks on ConFiguRe are challenging with a lot of room to improve. 
\end{itemize}

\section{Related Work}

Over the last decade, automated detection of figurative languages has become a popular topic, and a considerable number of datasets in this area have been constructed. These datasets can be roughly divided into two categories: the first is to extract the span of a specific figure given a sentence and its context~(span extraction); the second is to determine whether a sentence is figurative~(sentence classification). For span extraction, it is usually accomplished by marking out tokens carrying the target figure. These works include VUA~\cite{joshi_harnessing_2016} and the NTU Irony Corpus~\citep{tang_chinese_2014}. For sentence classification, there are two lines of relevant research. The first is a binary sentence classification task for determining whether a given sentence is figurative, e.g. SARC~\cite{reddit} for sarcasm, and the Chinese rhetoric question dataset built by \citet{rheq_dataset}. 
The second is to classify a figurative sentence into its corresponding figure type~(multi-classification), such as the Chinese multi-label rhetoric dataset constructed by \citet{chen_jointly_2021} for joint rhetoric and emotion identification.

On the basis of existing datasets, a series of methods have been developed, ranging from feature engineering~\cite{bulat-etal-2017-modelling, koper-schulte-im-walde-2017-improving, tsvetkov-etal-2014-metaphor} to neural networks~\cite{ liu_neural_2018, chen_jointly_2021, mu-etal-2019-learning, dankers-etal-2020-neighbourly, joshi_automatic_2017, leong_report_2020,leong_report_2018}. However, perhaps limited by the fact that existing datasets only target at one figure, previous works mainly deal with specific figure types. A comprehensive study of a collection of figure types remain under-explored.

One further observation is that previous datasets are rarely shipped with wider contextual information. They mainly approach the subject from a token-level or sentence-level perspective. 
That being the situation, a series of works have pointed out that leveraging contextual information is beneficial in figure detection~\cite{dankers-etal-2020-neighbourly, mu-etal-2019-learning, jang-etal-2015-metaphor, joshi_automatic_2017, joshi_harnessing_2015}. \citet{joshi_harnessing_2015} proposed to use text incongruity from linguistic theory for sarcasm detection.
\citet{dankers-etal-2020-neighbourly} used a general and a hierarchical attention mechanism for modeling discourse, improving SOTA for the 2018 VU Amsterdam (VUA) metaphor identification shared task~\citep{leong_report_2018} by 6.4 F1-scores. Other contextual clues such as author context~\cite{bamman_contextualized_2015, ghosh_magnets_2017}, multi-modal context~\cite{schifanella_detecting_2016,castro_towards_2019}, conversation context~\cite{joshi_harnessing_2016, ghosh_role_2017}, have also been proven to improve figure detection.

In this paper, we propose Chinese Corpus for Contextualized Figure Recognition~(ConFiguRe), which, to the best of our knowledge, is the first comprehensive corpus that includes more than 10 commonly used figures in Chinese, with relevant discourse fragments serving as contextual information for each instance. Our dataset can be used for both figure extraction and figure type classification.

\section{Corpus Construction}
Previous rhetorical studies \cite{oro15225, harris_annotation_2018} have contributed many useful annotation paradigms. However, 
These annotation paradigms are primarily customized for one specific figure type, or tailored to scheme figures that present themselves with strict patterns, which are quite inapplicable in our setting. In this section, we present our self-devised annotation guideline to build ConFiguRe with 7 trope figures and 5 scheme figures.

\subsection{Annotation Guideline}
\label{sec:anno_guide}
\begin{figure}
    \centering
    \includegraphics[scale=0.53]{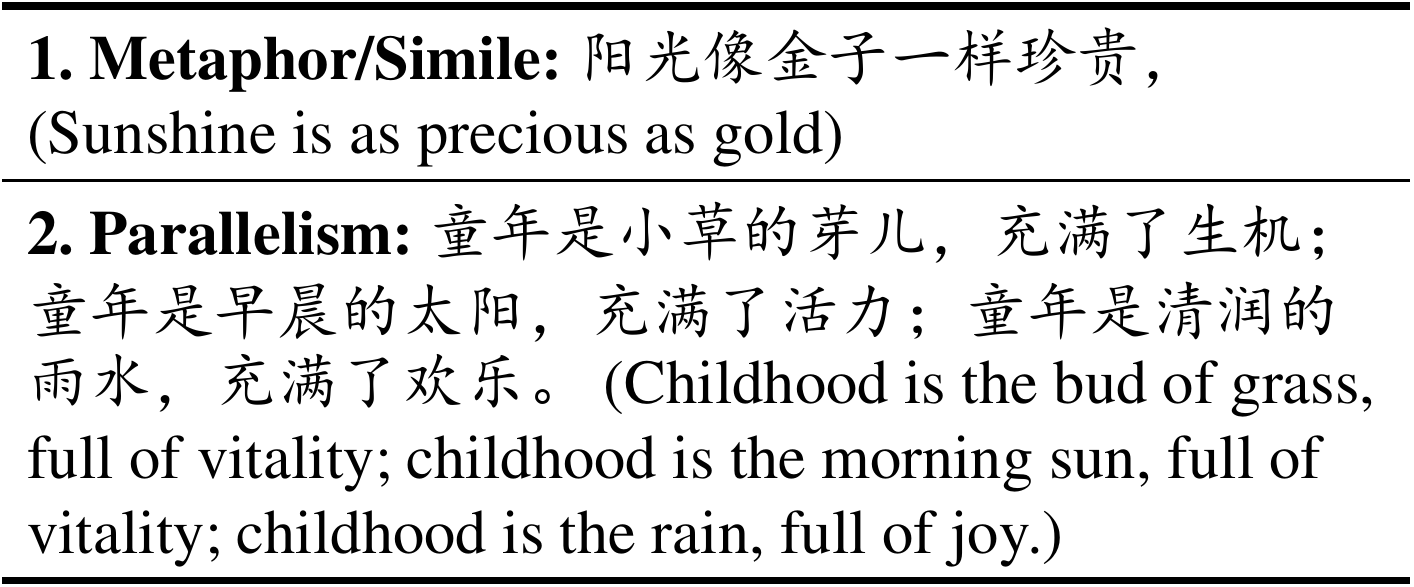}
    \caption{Two examples for figurative units. First is a \textit{Simile} unit containing one single clause, the second is a \textit{Parallelism} unit composed of six clauses.}
    \label{fig:fig_unit_example}
\end{figure}

\noindent \textbf{Figurative Unit}\quad For the convenience of analyzing figures, we first posit the concept of \textit{figurative unit} as the basic language carrier of a figure. A \textit{figurative unit} is defined as "the smallest continuous clause sequence carrying a complete expression of a specific figure". \footnote{It is debatable about the definition of a Chinese clause. Here for simplicity, we define a clause as a text span separated by the separator punctuations. A full list of punctuations is included in Appendix~\ref{app:punc}.}
The intuition is that, under most circumstances, a figure instance only comprises a limited portion of a sentence. Hence, we prefer clauses as elementary constituent of a figure, as it is more fine-grained than a sentence.
In Figure~\ref{fig:fig_unit_example}, we demonstrate two examples of figurative units with their corresponding figure types. In the first example, a single clause is a figurative unit carrying the figure \textit{Simile}; while in the second example, six clauses together form a figurative unit of \textit{Parallelism}.

\noindent \textbf{Figure Types}\quad
Upon selection of figure types, we refer to the linguistic categories from English~\cite{byu}  and  Chinese~\cite{Zhang63}. We choose the most frequently used ones in written literature. It should be noted that while some figures are widely used in English, they do not have exact counterparts in Chinese, therefore excluded from our dataset. At the same time, we avoid choosing the figures   whose identification may involve deep semantic background, e.g. Pun, Paradox, and leave them for future study. Specifically, we adopt the following
12 figures:
\textit{Metaphor/Simile}\footnote{In Chinese, these two figures are combined as a single figure type which means a comparison made by referring to one thing as another.}, 
\textit{Personification}, 
\textit{Metonymy}, 
\textit{Hyperbole}, 
\textit{Irony}, 
\textit{Synaesthesia}, 
\textit{Rhetorical question}, 
\textit{Parallelism}, 
\textit{Duality}, 
\textit{Repetition}, 
\textit{Antithesis}, 
\textit{Quote}~(For detailed description of each figure, see Appendix~\ref{app:fig_def}).
For simplicity, we temporarily use the first four or five characters as the abbreviation of each figure in this paper. Note that we do not aim at including all figure types in our research, but hope to recognize the commonly-used figures out of the discourse.

In linguistics, figures are divided into two main groups: \textit{schemes} and  \textit{tropes}. \textit{Schemes} reflect a deviation from the ordinary pattern or arrangement of words, while \textit{tropes} involve deviation from the ordinary and principal signification of words~\cite{corbett1999classical}.
Among the 12 commonly used figures, the first 7 are \textit{tropes} and the last 5 are \textit{schemes}. We adopt this categorization to facilitate following analysis below.

\subsection{Data Collection}
Though a common technique in communication, figurative languages are actually sparsely  distributed in main-stream corpora, consisting of news, dialogs, etc. For a higher proportion of figurative languages, we build our dataset on a collection of literary works. 98 Chinese literary works, most of which are novels and proses, are collected from publicly available resources.

We then divide each literary work into fragments comprising of whole paragraphs, while also ensuring that each fragment contains no more than 500 words. In this way, we assert that the discourse is neither too short to miss important contextual information nor too long to distract annotators' attention. In total, we obtain 12,976 fragments. These fragments are later presented to human annotators for manual annotation.

\subsection{Human Annotation}
We recruit 17 annotators whose native language is Chinese. These annotators are all well-educated, mostly majoring in linguistics. We divide them into two groups as \textit{fragment annotators} and \textit{figurative unit annotators}, provide instructions with detailed definitions and examples of each figure type, and train them for figure annotation before setting out on the full dataset. The annotation process is carried out coarse-to-fine in two stages.

The first stage is for coarse classification, where 5 fragment annotators are asked to classify fragments as figurative or not. Each fragment is annotated by one annotator only once, since we suppose classifying a discourse fragment as figurative or not is a binary classification task that is relatively easy.

The second stage involves 12 figurative unit annotators. Each figurative fragment is presented to two annotators, who are independently required to extract figurative units (one fragment can contain several units) and assign the corresponding figure types. We make sure that each figurative unit can only be labeled with one figure type. Additionally, conflict-solving strategies are devised in case of inconsistent labeling. For example, suppose annotators $A_1$ and $A_2$ respectively extract figurative units $u_1$ and $u_2$, the inconsistency can be roughly divided into 3 categories: 1) $u_1$ and $u_2$ are figurative units that do not overlap, in this case we keep both of them. 2) $u_1$ is a subset of $u_2$, and they are assigned the same figure type, in this case we choose $u_1$ as the gold annotation, since figurative unit is defined to be the smallest clause group containing a complete expression of a specific figure. 3) $u_1$ overlaps with $u_2$ but is neither a superset or subset, or they are assigned with different figure types, we regard this case as the most complex one and ask another annotator to make a decision based on the annotation results of  annotators $A_1$ and $A_2$. Besides, if an annotator extracts several figurative units in one go, inconsistency can be solved similarly.

In addition to identifying the figurative units and their types, we also ask the annotators to corroborate their judgement with evidence~(e.g. strong feature words denoting the figure) which may benefit future work. 

The main reason of designing this coarse-to-fine annotation process is that the work of fragment annotators can narrow down the context of a figure and make the figurative unit annotators pay more attention to identifying the boundary and type of a figure.

\subsection{Dataset Analysis}
\begin{table}[t]
    \centering
    \footnotesize
    \setlength\tabcolsep{3.5pt}
    \renewcommand{\arraystretch}{1.1}
    \begin{tabular}{cccccc}
         \toprule
         Split & \# frag. & wd./frag. & cls./frag. & \# figUnit. & figUnit./frag. \\
         \midrule
         train & 2934 & 419.7 &  41.9 & 6254 & 2.1\\
         valid & 419 & 417.5 & 42.1 & 886 & 2.1\\
         test & 839 & 419.2 & 41.3 & 1870 & 2.2\\ \midrule
         Total & 4192 & 419.4 &  41.8 &  9010&  2.1\\
         \bottomrule
    \end{tabular}
    \caption{Statistics of train, valid and test set in our ConFiguRe. \textit{frag.}, \textit{wd.}, \textit{cls.}, \textit{figUnit.} is short for fragment, word, clause, figurative unit, repectively.}
    \label{tab:fig_split}
\end{table}

\begin{table}[t]
    \centering
    \footnotesize
    \setlength\tabcolsep{5pt}
    \renewcommand{\arraystretch}{1.1}
    \begin{tabular}{cccccc}
         \toprule
         \multirow{2}{*}{Figures} & 
         \multicolumn{3}{c}{\# figurative unit} & 
         \multirow{2}{*}{words/unit} & 
         \multirow{2}{*}{clauses/unit} \\
         \cmidrule(r){2-4} & train & valid & test & &  \\ 
         \midrule
         Meta. & 2226 & 306 & 683 & 18.5 & 1.55  \\
         Pers. & 768 & 94 & 243 & 19.5 &  1.64 \\
         Meto. & 411 & 69 & 123 & 18.3 &  1.68\\ 
         Hyper. & 473 & 72 & 145 & 19.5 & 1.75 \\ 
         Irony & 24 & 4 & 5 & 27.5 & 2.36\\ 
         Syna. & 18 & 6 & 10 & 22.9 & 1.76  \\ 
         Rheq. & 868 & 115 & 202 & 19.5 & 2.07 \\ \midrule
         Para. & 291 & 41 & 99 & 37.1 & 3.90  \\ 
         Dual. & 272 & 37 & 76 & 16.8 & 2.24  \\ \
         Repe. & 382 & 57 & 105 & 16.6 & 2.98 \\ 
         Anti. & 179 & 23 & 62 & 28.7 & 2.73  \\ 
         Quote & 342 & 63 & 117 & 33.3&  3.90 \\ \midrule
         Total & 6254 & 886 & 1870 & 20.7 & 2.05  \\
         \bottomrule
    \end{tabular}
    \caption{Statistics of each figure in train, valid and test set. For simplicity, we temporarily use the first four or five characters as the abbreviation of each figure in this paper. First seven figures are \textit{tropes}, last five figures are \textit{schemes}.}
    \label{tab:fig_data}
\end{table}

Through human annotation, we present ConFiguRe with 4,192 figurative fragments and 9,010 figurative units. Note that each instance in ConFiguRe is a discourse fragment carrying figurative languages. Each fragment may contain one or more annotated figurative units. Table~\ref{tab:fig_split} demonstrates basic statistics of the  train, valid and test set, according to the split proportion of 7:1:2. 

Detailed information of each figure type is rendered in Table~\ref{tab:fig_data}. We can see that \textit{metaphor/simile}, \textit{personification}, and \textit{rhetoric question} occur more frequently than others, while \textit{synaesthesia} and \textit{irony} are extremely hard to gather in our corpus. For these two figure types, we may collect more instances in the future.
From Table ~\ref{tab:fig_data}, we can also see that the average number of clauses for \textit{schemes} is consistently larger than that of \textit{tropes}. This can be attributed to the fact that \textit{schemes} are usually only identifiable in a sequence of clauses, while \textit{tropes} can be directly expressed within one clause. Interestingly, the average word number of \textit{irony} figures is comparable to or even longer than many  scheme figures. A priori is that the expression of an \textit{irony} requires more  contextual information. Scheme figures \textit{duality} and \textit{repetition} are relatively short, probably due to the fact that \textit{duality} often leverages short clauses and \textit{repetition} evinces at the word level. 

\section{Task Definition and Baseline Models}
Based on ConFiguRe, we propose three benchmark tasks. Task 1 is figure extraction, which extracts figurative units from input text. Task 2 is figure type classification, which classifies a figurative unit into the corresponding figure type. Task 3 is figure recognition composed of previous two tasks, which extracts figurative units from input text and determines their corresponding figure types simultaneously. 

Although there have been a lot of methods in rhetoric detection, we find that these methods mostly catered for one specific figure~\cite{leong_report_2020,ghosh_report_2020}, and therefore not quite consistent with our task settings. For baseline models, we adopt self-designed approaches backboned with the state-of-the-art RoBERTa model.
We give a formal definition for each task and introduce the corresponding baseline models as follows.

\subsection{Task 1: Figure Extraction}

Given a discourse fragment $\mathcal{D}$ comprising $m$ clauses and $n$ words, let $\mathcal{D} = \left\{ c_i \right\}_{i=1}^{m} $, where $c_i = \left\{w_j\right\}_{j=1}^{n_i}$ refers to the $i$-th clause consisting of $n_i$ words, and $\sum_{i=1}^{m}n_i=n$.
Task 1 aims to extract figurative units $u_1=(c_{b_1}, \dots, c_{e_1}), \cdots, u_k=(c_{b_k}, \dots, c_{e_k})$ from $\mathcal{D}$, where $k$ is the number of figurative units in $\mathcal{D}$, $u_i$ is the $i$-th figurative unit, and the subscripts $b_i$ and $e_i$ denote the beginning and ending positions, respectively. Note that each figurative unit is the smallest continuous clause sequence carrying a complete expression of a specific figure. (See Section~\ref{sec:anno_guide} for detailed explanation of clause and figurative unit)

To perform figure extraction, we first use RoBERTa encoder and perform clause-wise mean pooling to obtain contextualized representations $\boldsymbol{h_c} = (h_{clause_1},\dots, h_{clause_m})$ for each clause, where $m$ is the number of clauses. Based on this, we design the following baselines for Task 1:

 \noindent \textbf{FESeq}\quad FESeq is implemented by modeling figure extraction as a clause-level sequence labeling task. Following traditional settings, the labels "B" "I" and "O" are assigned to each clause when it is the first clause of a figurative unit, inside but not the first of a figurative unit, and not in a figurative unit accordingly. A classification layer is added on top of RoBERTa, casting hidden representations into 3-dimensional logits for "B", "I" or "O".

 \noindent \textbf{FECRF}\quad Since Conditional Random Field~\cite{10.5555/645530.655813} is a common strategy in sequence labeling tasks, we design the FECRF model by adding a CRF layer on top of FESeq model.

 \noindent \textbf{FESpan}\quad FESpan is developed by modeling figure extraction as a clause-level binary span tagging task.
It adopts two binary MLP classifiers upon encoder to detect the start and end  position for each figurative unit, respectively.
More precisely, FESpan assigns each clause a binary tag (0/1), which indicates whether the current clause corresponds to a start or end position of a figurative unit.

\subsection{Task 2: Figure Type Classification}

Given a figurative unit comprising $n$ words
$ \boldsymbol{u}= \left\{w_i \right\}_{1}^{n}$, and its context comprising $m$ words $\mathcal{C} =\left\{w_j \right\}_{1}^{m}$, task 2 aims to classify this unit $\boldsymbol{u}$ 
into the right figure type.

First, the RoBERTa encoder produces contextualized representations ${h_i}$ for each token $w_i$:
\begin{gather}
    \boldsymbol{h} = (h_1, \dots, h_n) = \text{Encoder}(w_1, \dots, w_n)
    \label{eqn: token_h}
\end{gather}
Mean pooling is then applied on $\boldsymbol{h}$ to get hidden representation
$h^{\boldsymbol{u}} $ 
for the figure unit
$\boldsymbol{u}$.
Based on this, we design the following baselines for task 2:

\noindent
\textbf{FTCLS}\quad FTCLS is built by adding a classification head on top of RoBERTa, which then classifies the input into one of 12 figure types. 

\noindent
 \textbf{FTCXT}\quad Since context information has proven to yield improvement for metaphor and sarcasm detection~\cite{dankers-etal-2020-neighbourly, mu-etal-2019-learning, joshi_harnessing_2016, ghosh_role_2017}, FTCXT is designed to incorporate contextual information in figure classification and explore its effect across all figure types. Specifically, hidden representation $h^{\mathcal{C}}$ for context $\mathcal{C}$ is also calculated by the encoder. $h^{\boldsymbol{u}}$ 
 and $h^{\mathcal{C}}$ are then concatenated  for classification.
\begin{gather}
    f = \text{Classification}(\left[ h^{\boldsymbol{u}};h^{\mathcal{C}} \right])
\end{gather}
where $f$ is the predicted figure type.

\subsection{Task 3: Figure Recognition}

Similar to task 1, given a discourse fragment $\mathcal{D}$ comprising $m$ clauses and $n$ words, 
Task 3 aims to extract figurative units 
$(u_1, \cdots, u_k)$
from $\mathcal{D}$, and classify each figurative unit into a specific figure type as $f_1, \cdots, f_k$ in the same time, where $k$ is the number of figurative units in text $\mathcal{D}$. 

Following task 1, contextualized representations $\boldsymbol{h_c} = (h_{c_1},\dots, h_{c_m})$ for the $m$ clause are produced by RoBERTa encoder.
Based on this, we define following baselines:

\noindent \textbf{Rule-based Method}\quad Some figures in our dataset manifest obvious patterns. For example,
the type of \textit{Metaphor/Simile}
usually comes with indicators such as the Chinese words \begin{CJK*}{UTF8}{gbsn}``像~(like)'' \end{CJK*},  \begin{CJK*}{UTF8}{gbsn}``如~(as)''\end{CJK*}.
The \textit{Parallelism} type is composed of similar clauses, most commonly separated by colon. To exploit these obvious patterns, we design heuristic rules for figure recognition as a complement to our neural methods.

\noindent
\textbf{Pipeline}\quad Since figure recognition can be naturally tackled as first extraction and then classification, we set our pipeline baseline as the combination of best-performing models in extracion task and classification task. Specifically, we first use FECRF to extract figurative units from input text, then use FTCXT to classify the extracted figurative units into their corresponding figure types.

\noindent
 \textbf{E2ESeq}\quad By modeling figure recognition as a sequence labeling task, E2ESeq model shares the same architecture as the FESeq model mentioned before. The difference is that, in this case, we assign clauses of different figure types with different "B" and "I" labels. For instance, for a figurative unit of irony, we assign "B-Irony" to its first clause and "I-Irony" to other clauses in this unit.

\noindent
 \textbf{E2ECRF}\quad Similar to FECRF, we add a CRF layer on top of the above E2ESeq model to construct a E2ECRF model.
\section{Experiments}
\begin{table*}[t]
    \centering
    \footnotesize
    \setlength\tabcolsep{13pt}
    \renewcommand{\arraystretch}{1.1}
    \begin{tabular}{cccccc}
         \toprule
         Task & Model & Precision & Recall & Micro F1 & Macro F1 \\ \midrule
         \multirow{3}{*}{Extraction} 
         & FESeq & 34.06 & 29.23 & 31.46 & - \\
         & FECRF & 32.56 &  \textbf{31.60} & \textbf{32.07} & - \\
         & FESpan & \textbf{39.41} & 24.61 & 30.30 & - \\
         \midrule 
         \multirow{2}{*}{Classification} & FTCLS & 78.82 & 78.82 & 78.82 & 65.22 \\
         & FTCXT & \textbf{79.49} & \textbf{79.49} & \textbf{79.49} & \textbf{68.72} \\ 
         \midrule
         \multirow{4}{*}{Recognition} 
         & Rule & 7.74 & 12.76 & 9.64 & 5.24 \\
         & Pipeline & \textbf{31.10} & 25.52 & 28.03 & 18.17 \\
         & E2ESeq & 29.21 & 25.97 & 29.47 & 17.35 \\
         & E2ECRF & 30.87 & \textbf{31.10} & \textbf{30.99} & \textbf{21.27} \\
         \bottomrule
    \end{tabular}
    \caption{Main experiment results. We highlight the highest numbers among models in  \textbf{bold}.}
    \label{tab:main}
\end{table*}
\noindent \textbf{Implementation Details}\quad We implement our models using HuggingFace's Transformers~\citep{wolf-etal-2020-transformers}. We choose the RoBERTa-zh-Large~\cite{cui-etal-2020-revisiting} checkpoint
trained on Chinese corpus.
For fine-tuning, we generally stick to a dropout rate of 0.1, a batch size of 16, an epoch of 30, the Adam~\cite{kingma2017adam} optimizer, and a learning rate of $1e^{-5}$.
We select our hyperparameters based on the best performance on validation set and report the average results from 5 runs with different random seeds.

We present main experiment results on our dataset for all three tasks in Table~\ref{tab:main}. For the extraction task, we report precision, recall and Micro F1 score for each model. For other tasks, we additionally report Macro F1 score by averaging out F1 scores of all figure types. We also conduct subsidiary experiments and analysis for each task. By doing so, we shed light on promising directions for future work.

\subsection{Evaluating Figure Extraction}
\label{sec5_1}

Model performance for figure extraction is presented in the first block of Table~\ref{tab:main}. FECRF yields the best result, surpassing FESeq by 1.27 F1 score and FESpan by 1.77 F1 score, suggesting that figure extraction can be modeled as a sequence labeling problem which CRF is specialized in.
We also find that FESpan gives the highest precision but low recall value. This may be attributed to the fact that using two classifiers to predict the boundaries of figures imposes further restrictions.

\begin{table}[t]
    \centering
    \footnotesize
    \setlength\tabcolsep{8pt}
    \renewcommand{\arraystretch}{1.1}
    \begin{tabular}{ccc}
         \toprule
          \multirow{2.5}{*}{Figure} &  Extraction & Recognition \\
          & (FESeq) & (E2ECRF) \\
         \midrule
         Exact match & 572 & 563\\
         Wrong figure type & - & 75\\ \midrule
         Super prediction & 201 & 215\\
         Sub prediction & 289 & 227\\
         Overlapping prediction & 40 & 45\\
         Non-lapping prediction & 665 & 699 \\ \midrule
         Total predictions & 1757 & 1824\\
         \bottomrule
    \end{tabular}
    \caption{Error analysis for model predictions in extraction and recognition task.}
    \label{tab:error_ana}
\end{table}

Overall, the performance of figure unit extraction
is not satisfactory, since it is somewhat difficult to precisely compartmentalize the smallest clause sequence containing figurative languages. Only 572 out 1757 predictions exactly matches gold label.
We observe that the prediction errors can be largely categorized as follows:
1) \textit{Super/Sub prediction}, the predicted clause sequence is a superset/subset of the gold figurative unit; 
2) \textit{Overlapping prediction}, the predicted clause sequence overlaps with a part of the gold figurative unit but is neither its superset nor subset; 
3) \textit{Non-lapping prediction}, the predicted clause sequence does not overlap with any gold figurative unit, i.e. predicting non-figurative clause groups as figurative. 
Column 2 in Table \ref{tab:error_ana} presents the numbers of each error category according to prediction results of FESeq.
More than half of the wrong predictions do not overlap with any gold figurative unit. 
Other wrong predictions are mainly supersets or subsets of certain gold figurative unit. 
Based on these observations, we conclude that the SOTA models are not doing very well in either discerning figurative languages or determining the exact boundary of figurative unit in our corpus, perhaps because the size of our corpus is relatively small and imbalanced.

\subsection{Evaluating Figure Type Classification}

For figure type classification, we present 
results of FTCLS and FTCXT in the second block of Table~\ref{tab:main}. The latter gives an improvement of 0.67 micro F1 point and 3.5 macro F1 point over the former.
Compared to figure extraction, the performance of figure type classification exhibits a high score.

\begin{table}[t]
    \centering
    \footnotesize
    \setlength\tabcolsep{5pt}
    \renewcommand{\arraystretch}{1.1}
    \begin{tabular}{ccccccc}
         \toprule
         \multirow{2.5}{*}{Figure} & \multicolumn{3}{c}{FTCLS} &
         \multicolumn{3}{c}{FTCXT} \\
         \cmidrule(r){2-4} \cmidrule(r){5-7}  & P & R & F1 & P & R & F1 \\ 
         \midrule
         Meta. & 84.16 & 87.12 & 85.61 & 83.40 & 88.29 & 85.78 \\
         Pers. & 66.80 & 67.08 & 66.94 & 66.27 & 67.99 & 67.07 \\
         Meto. & 62.24 & 49.59 & 55.20 & 66.67 & 47.15 & 55.24 \\
         Hyper. & 57.66 & 54.48 & 56.03 & 58.06 & 49.66 & 53.53 \\
         Irony & 50.00 & 20.00 & 28.57 & 50.00 & 40.00 & 44.44 \\
         Syna. & 0.00 & 0.00 & 0.00 & 25.00 & 10.00 & 14.29 \\
         Rheq. & 93.75 & 96.53 & 95.12 & 92.89 & 97.03 & 94.92 \\
         \midrule
         Para. & 79.66 & 94.95 & 86.64 & 82.05 & 96.97 & 88.89 \\
         Dual. & 70.13 & 71.05 & 70.59 & 70.13 & 71.05 & 70.59 \\
         Repe. & 89.91 & 93.33 & 91.59 & 90.65 & 92.38 & 91.51 \\
         Anti. & 67.31 & 56.45 & 61.40 & 78.43 & 64.52 & 70.80 \\
         Quote & 85.34 & 84.62 & 84.98 & 87.93 & 87.18 & 87.55 \\
         
         \bottomrule
    \end{tabular}
    \caption{Figure type classification results w/ or w/o contextual information for each figure of speech. First seven figures are \textit{tropes}, next five figures are \textit{schemes}.}
    \label{tab:class_result}
\end{table}

Detailed classification results for each figure w/o contextual information is presented in Table \ref{tab:class_result}. 
From this table, Columns 2-4 
give FTCLS' classification results for each figure. It can be seen that classification accuracy is quite imbalanced across all figure types. On the one hand, F1 scores on \textit{schemes} are relatively higher than \textit{tropes} on the whole. We suppose it is because \textit{schemes} tend to manifest obvious patterns like repetition, which is easier for a model to capture, while \textit{tropes} involve deviation from superficial meaning, which is more challenging to models and even humans. We expect more efforts on modeling semantics to benefit figure type classification. Notably, \textit{Rhetoric Question~(Rheq.)} gives the highest F1 score of 95.12 while being a trope. We suppose it is because the \textit{question mark~(i.e. "?")} in \textit{Rheq.} units serve as an obvious clue to facilitate classification. 

Table \ref{tab:class_result} reveals that
even among trope figures, model performance is quite imbalanced. F1 score on Metahpor achieve 85.61 while on irony and synaesthesia it is 28.57 and 0.00, repectively. We suppose the reason for this performance gap is twofold. Firstly, figures such as irony and synaesthesia are demanding to collect, resulting in their fairly low distribution in ConFiguRe. To effectively train models for such low-frequency figures, it is necessary to incorporate other specialized techniques. 
Secondly, figures as epitomized by irony, are strongly related to a wider context, without which the identification of such figures becomes insufficient.

\noindent \textbf{Context Benefits Figure Type Classification}\quad 
To inspect the extent that context information benefits figure classification, we include result of FTCXT in Columns 5-7  of Table \ref{tab:class_result}. This serves as an "ablation study" that undergirds the usefulness of contextual information in figure classification.
By comparing results of FTCLS and FTCXT, it can be seen that context information consistently improves classification accuracy for most figures. 
Further, among all figure types, contextual information is especially conducive to Irony, Synaesthesia and Antithesis, respectively boosting F1 score by 15.87, 14.29 and 9.40 point. 
This observation is consistent with the linguistic assumption that the identification of these figures usually depends on more context. 

\subsection{Evaluating Figure Recognition}

For figure recognition, baseline results can be found in the third block of Table~\ref{tab:main}. Similar to figure extraction task, we observe that the sequence labeling model with CRF achieves the best result, surpassing the non-CRF version with 1.52 micro F1 score and 3.92 macro F1 score. It also outperforms the pipeline approach combining the best-performing extraction model and classification model, which suffers from error propagation. F1 score of the rule-based method is the lowest compared to pipeline and end2end methods, indicating that figure recognition requires much more efforts beyond recognizing shallow and obvious patterns. 

Same as figure extraction task, errors of this task incorporates the following types: \textit{Super prediction}, \textit{Sub prediction} and
\textit{Overlapping prediction}. 
Besides,
we introduce \textit{Wrong figure type} error, 
where the model successfully extracts the figurative units but classifies them into wrong figure type. According to the result of E2ECRF, numbers of each error type is presented in  Column 3 of Table \ref{tab:error_ana}. We observe that a high percentage of precisely identified figure units are correctly classified with the figure type, and only  75 out of 638 predictions fail in classification.
Resonating with figure extraction, we conclude that \textbf{discerning figurative units is quite challenging for state-of-the-art models. }

\subsection{Discussion}
\begin{figure}
    \centering
    \includegraphics[scale=0.53]{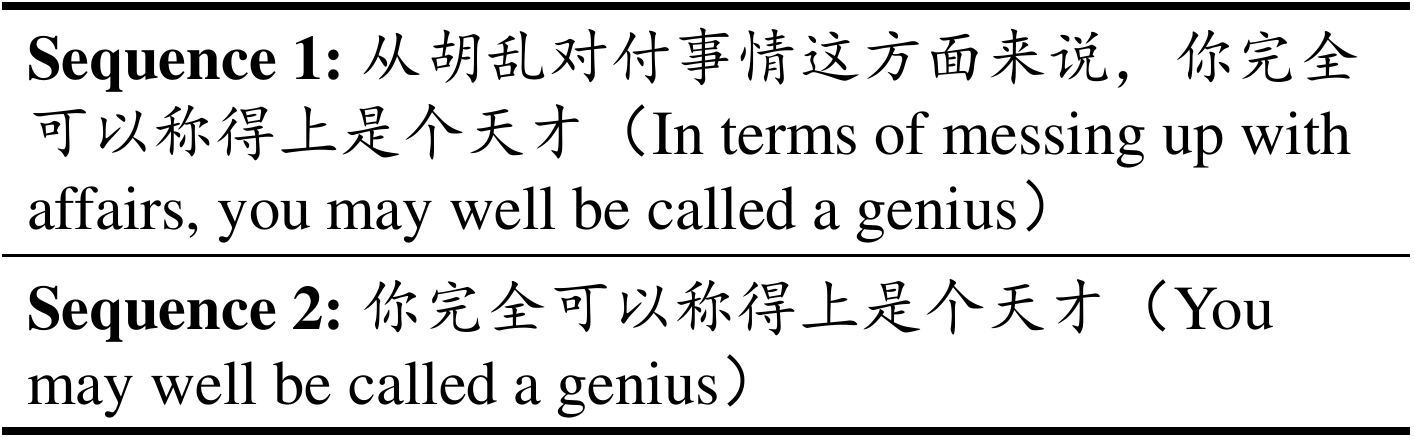}
    \caption{Example of two clause sequences marked as figurative unit for \textit{irony}.}
    \label{fig:case_study}
\end{figure}
\noindent \textbf{Revisiting Figurative Unit}\quad
From Subsection \ref{sec5_1}, we obtain a rather low performance in recognizing figurative units. We investigate the classification results of the models and find them rather satisfying. We therefore impute the poor performances to the difficulties in delimiting the boundary. Even if we clearly define a figure unit as the smallest clause sequence, such concept is somewhat controversial under certain circumstances.
Figure~\ref{fig:case_study} illustrates two instances which are ambiguous in deciding the gold figure unit.
For this reason, it is necessary to improve the annotation process and evaluation metrics in the future.

\noindent\textbf{Revisiting Tropes and Schemes}\quad
\begin{figure}
    \centering
    \includegraphics[scale=0.45]{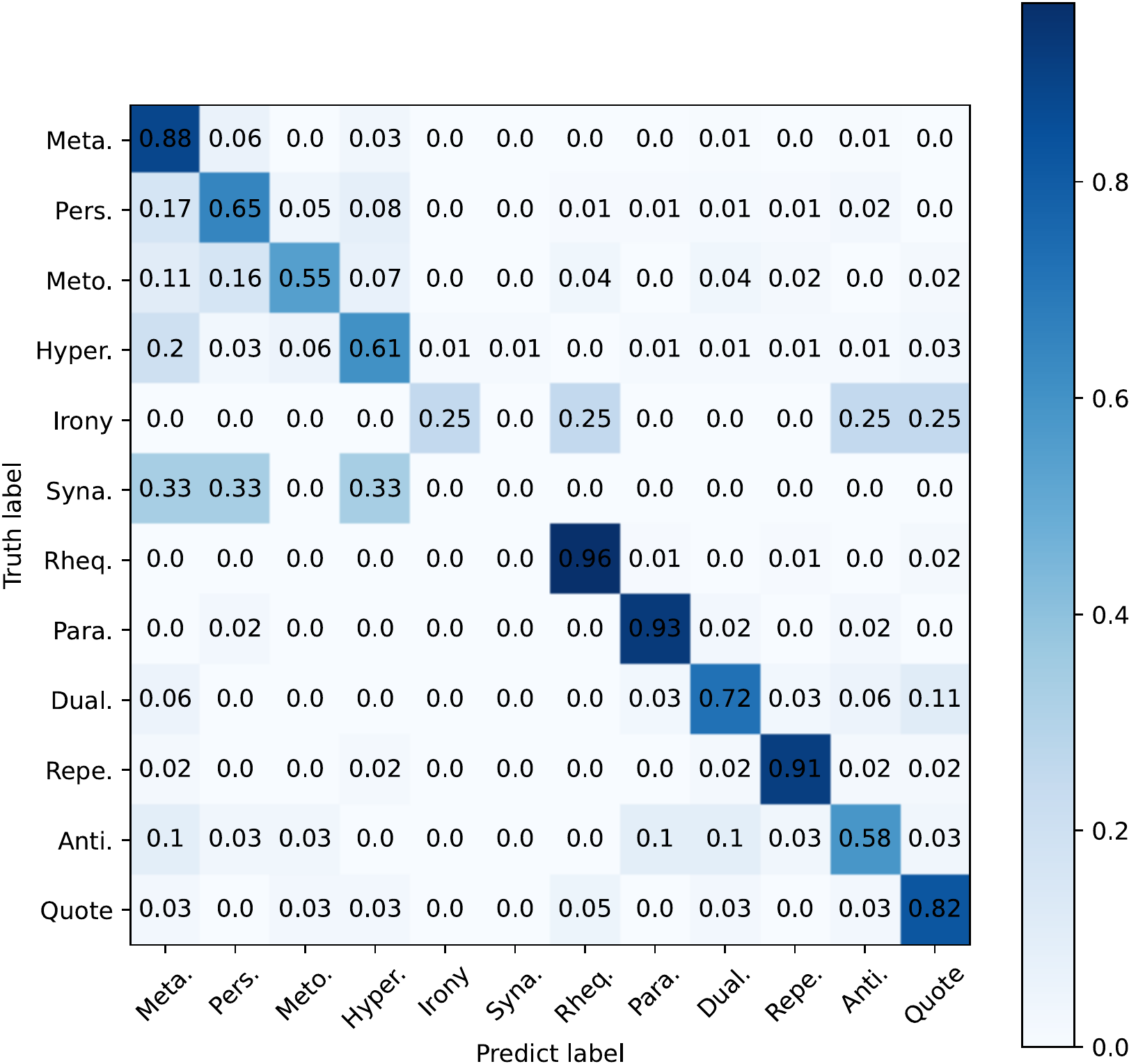}
    \caption{Confusion matrix from FTCLS's results in figure type classification. First seven figures are \textit{tropes}, last five figures are \textit{schemes}.}
    \label{fig:confusion_matrix}
\end{figure}
Interestingly, our work can also serve as a supporting evidence for the linguistic categories of \textit{tropes} and \textit{schemes}.  \textit{Schemes} reflect deviation from the ordinary pattern or arrangement of words, while \textit{tropes} involve deviation from the ordinary and principal signification of a word~\cite{corbett1999classical}. A heat map of confusion matrix in figure type classification is presented in Figure~\ref{fig:confusion_matrix}. It is intuitively suggested that, even misclassified, the predicted label tends to fall into the same category as the gold label. For the example of \textit{hyperbole}, there is a 61\% probability of being classified correctly, a  20\% probability as \textit{metaphor/simile},  and a  6\% probability as \textit{metonymy}, all falling into the category of \textit{tropes}. In the meantime, the probability of being classified as \textit{schemes} is merely 7\%. In other words, a trope figure is more similar to other trope figures than scheme figures, and vice versa. 
\section{Conclusions and Future Work}

In this paper, we argue that it is necessary to recognize figures from the discourse level.  
For the first time, we propose the concept of figurative unit as language carrier of one figure and construct a Chinese corpus for Contextualized Figure Recognition~(ConFiguRe).
ConFiguRe includes 12 most commonly used figures in Chinese, with discourse-level contextual information attached to each figure instance.
On ConFiguRe, we design three tasks of figure extraction, figure type classification and figure recognition, and implement state-of-the-art models. A series of experiments show that all three tasks remain challenging and worth exploring.

In future, we hope to increase the size of our corpus, especially adding more figure instances for the types with fewer instances such as irony and synaesthesia.
At the same time, we will to improve the model performance with consideration of incorporating contextual information and using more training data. 

\section*{Acknowledgement}

We thank the anonymous reviewers for their helpful comments on this paper. This work was partially supported by National Key Research and Development Program of China (2020AAA0109703),
National Natural Science Foundation of China (61876009), and National Social Science Foundation Project of China (21\&ZD287).

\bibliography{bib/anthology,bib/custom}
\bibliographystyle{bib/acl_natbib}


\appendix

\section{Annotation Details}
\label{sec:appendix}

\subsection{Figure Definition}
\label{app:fig_def}
\begin{figure*}
    \centering
    \includegraphics[scale=0.95]{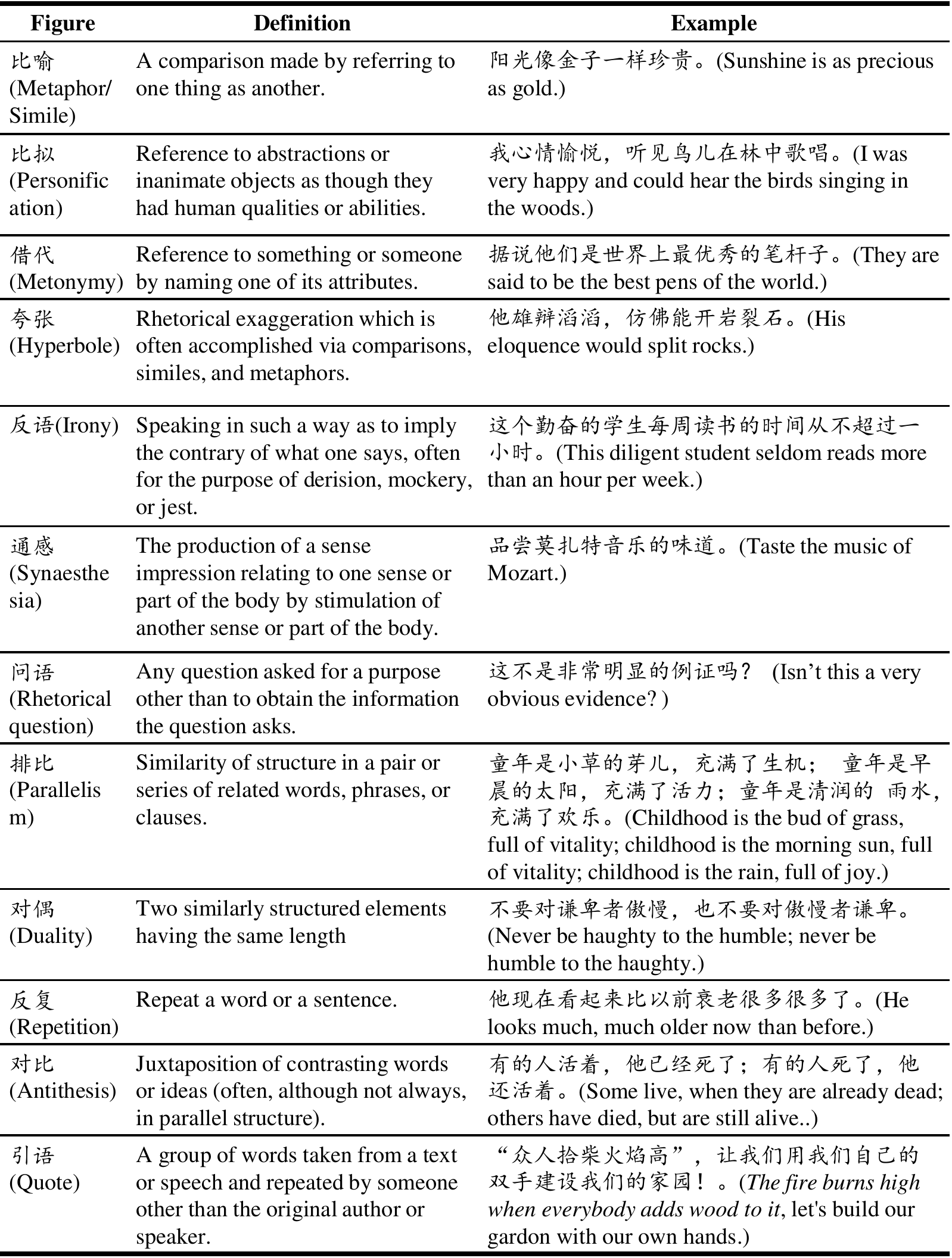}
    \caption{Definition and example for each figure type.}
    \label{fig:figure_def}
\end{figure*}

We define these 12 figure categorites in reference to silva rhetoricae~(\url{http://rhetoric.byu.edu/}), an authoritative website for rhetoric figures. We also follow a canonical work on rhetorics by Chinese linguistics Gong Zhang (1963). Detailed definition and example for each figure type above is provided in Figure~\ref{fig:figure_def}.

\subsection{Punctuation List}
\label{app:punc}

Below is punctuation list we use to separate clauses: 
\{
\begin{CJK*}{UTF8}{gbsn}\textbf{，} \end{CJK*},
\begin{CJK*}{UTF8}{gbsn}\textbf{；} \end{CJK*},
\begin{CJK*}{UTF8}{gbsn}\textbf{。} \end{CJK*}, 
\begin{CJK*}{UTF8}{gbsn}\textbf{？} \end{CJK*}, 
\begin{CJK*}{UTF8}{gbsn}\textbf{！} \end{CJK*}, 
\begin{CJK*}{UTF8}{gbsn}\textbf{……} \end{CJK*}, 
\begin{CJK*}{UTF8}{gbsn}\textbf{——} \end{CJK*},
\begin{CJK*}{UTF8}{gbsn}\textbf{：} \end{CJK*}
\}

\end{document}